\documentclass[cameraready]{Interspeech}

\usepackage{fontspec}
\usepackage{newtxtext} 
\usepackage{newtxmath}

\usepackage{tcolorbox}
\usepackage{xcolor}
\usepackage{soul} 
\usepackage{pifont} 
\usepackage[normalem]{ulem}
\usepackage{multirow}
\usepackage{multicol}
\tcbuselibrary{skins, breakable}

\definecolor{ErrorRed}{HTML}{E74C3C}
\definecolor{CorrectGreen}{HTML}{27AE60}
\definecolor{BoxGray}{HTML}{F8F9F9}
\definecolor{BorderBlue}{HTML}{2980B9}
\title{Interactive ASR: Towards Human-Like Interaction and Semantic Coherence Evaluation for Agentic Speech Recognition%
\thanks{Project page: \url{https://interactiveasr.github.io/}. Live demo: \url{https://i-asr.sjtuxlance.com/}}}
\author[affiliation={2},equalcontribution]{Peng}{Wang}
\author[affiliation={1},equalcontribution]{Yanqiao}{Zhu$^\dag$}
\author[affiliation={3},equalcontribution]{Zixuan}{Jiang}
\author[affiliation={4}]{Qinyuan}{Chen}
\author[affiliation={4}]{Xingjian}{Zhao}
\author[affiliation={4}]{Xipeng}{Qiu}
\author[affiliation={5}]{Wupeng}{Wang}
\author[affiliation={5}]{Zhifu}{Gao}
\author[affiliation={5}]{Xiangang}{Li}
\author[affiliation={1}]{Kai}{Yu}
\author[affiliation={1},correspondingauthor]{Xie}{Chen}

\address{
    $^1$ X-LANCE Lab, Shanghai Jiao Tong University   $^2$ The Chinese University of Hong Kong, Shenzhen\\
    $^3$ Xi'an Jiaotong University   $^4$ Fudan University\\
    $^5$ Tongyi Fun Team, Alibaba Group
}
\email{pengwang0104@gmail.com, 1850432206@sjtu.edu.cn, andrewjiang@stu.xjtu.edu.cn, chengqy21@m.fudan.edu.cn, zhaoxj24@m.fudan.edu.cn, xpqiu@fudan.edu.cn, wangwupeng.wwp@alibaba-inc.com, zhifu.gzf@alibaba-inc.com, lixiangang.lxg@alibaba-inc.com, kai.yu@sjtu.edu.cn, chenxie95@sjtu.edu.cn}

\keywords{
speech recognition,
human-computer interaction,
LLM agent
}

\usepackage{comment}

\begin{document}
\ninept

\maketitle
\def\thefootnote{\dag}
\footnotetext{This work was conducted during an internship at Tongyi Fun Team, Alibaba Group.}
\def\thefootnote{\arabic{footnote}}
\begin{abstract}
Recent years have witnessed remarkable progress in automatic speech recognition (ASR), driven by advances in model architectures and large-scale training data. However, two important aspects remain underexplored. First, Word Error Rate (WER), the dominant evaluation metric for decades, treats all words equally and often fails to reflect the semantic correctness of an utterance at the sentence level. Second, interactive correction—an essential component of human communication—has rarely been systematically studied in ASR research. In this paper, we integrate these two perspectives under an agentic framework for interactive ASR. We propose leveraging LLM-as-a-Judge as a semantic-aware evaluation metric to assess recognition quality beyond token-level accuracy. Furthermore, we design an LLM-driven agent framework to simulate human-like multi-turn interaction, enabling iterative refinement of recognition outputs through semantic feedback. Extensive experiments are conducted on standard benchmarks, including GigaSpeech (English), WenetSpeech (Chinese), the ASRU 2019 code-switching test set. Both objective and subjective evaluations demonstrate the effectiveness of the proposed framework in improving semantic fidelity and interactive correction capability. We will release the code to facilitate future research in interactive and agentic ASR.

\end{abstract}

\begin{figure}[t!]
    \centering
    \centerline{\includegraphics[width=5cm]{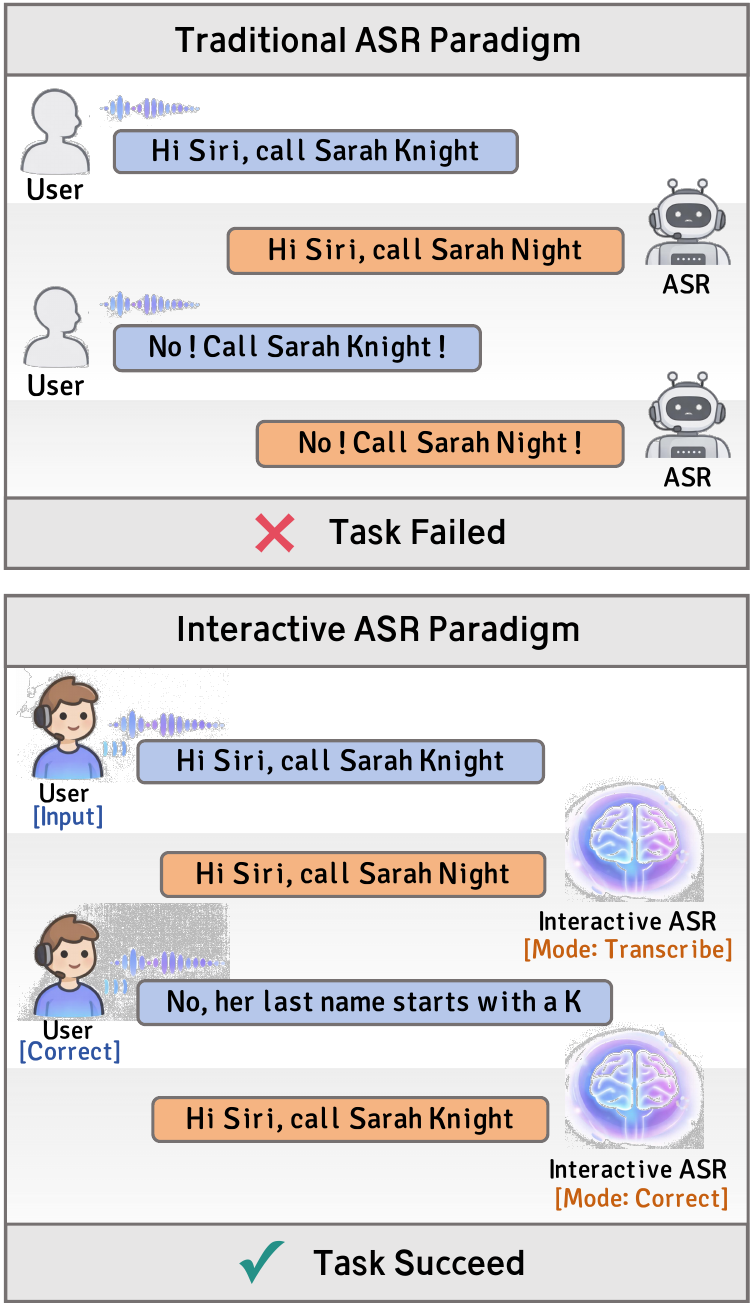}}
    \caption{
Traditional vs. Interactive ASR paradigms. (Top) Traditional systems struggle with named entities (e.g., "Night" vs. "Knight"); (Bottom), our proposed Interactive ASR can take user's spoken language instructions (e.g., "starts with a K") as the feedback to update and correct ASR results.
}
    \label{fig:Traditional asr paragidm vs Interactive ASR Paradigm.}
    \vspace{-0.7cm}
\end{figure}

\section{Introduction}

Automatic speech recognition (ASR) plays a pivotal role in human–computer interaction by enabling computers to understand users’ intent through speech. In recent years, ASR technologies have achieved remarkable progress, driven by advances in both model architectures and large-scale training data. Extensive research has explored a variety of modeling paradigms, ranging from end-to-end approaches \cite{ctc_graves, rnnt_graves, aed_chan,aed_whisper_radford}, to more recent large language model (LLM)-based frameworks \cite{wu2023decoder,wang2023slm,tang2023salmonn,chu2023qwen,fathullah2024prompting,ma2024embarrassingly}. Meanwhile, scaling laws in both model capacity and training data have proven highly effective in further advancing ASR performance \cite{seedasr_bai,fireredasr_xu,funasr_an,qwen3asr_shi}. 

However, despite the rapid progress in ASR, several important aspects remain underexplored. The first concerns the evaluation metrics used for ASR systems. For decades, Word Error Rate (WER) has been widely adopted to measure the discrepancy between recognition hypotheses and reference transcriptions by aligning words in the two sequences. Despite its simplicity and objectivity, this de facto metric has several well-known limitations. In particular, WER treats all words equally and assigns the same penalty to each recognition error, while different types of words may affect sentence semantics differently \cite{roy2021semantic}. For example, errors in function words often have limited impact, whereas misrecognition of critical content words such as named entities can significantly alter the intended meaning. As a result, WER may fail to adequately reflect the semantic impact of recognition errors. Meanwhile, large language models (LLMs) have demonstrated strong semantic understanding capabilities and can often infer or correct minor recognition errors during interaction\cite{liu2025denoising}. This suggests that future evaluation of ASR systems should move beyond word-level accuracy and focus more on errors that may mislead downstream language understanding.

The second issue lies in the integration of user feedback. In human communication, clarification and correction through interaction are common when key information is ambiguous or misunderstood. However, such interactive correction has been largely overlooked in human–computer interaction. Most current ASR systems cannot revise their recognition outputs once errors occur, even when users explicitly point them out, which may significantly degrade user experience, particularly in on-screen scenarios. Moreover, the inherent ambiguity of spoken language—such as homonyms in names or recognition errors caused by accents or background noise—further motivates the development of ASR systems that can interact with users to clarify and correct recognition results.

To address these challenges and adapt ASR systems to the LLM era, we propose \textbf{Interactive ASR}, a framework that integrates semantic-aware evaluation and user interaction. First, we revisit LLM-as-a-Judge\cite{zheng2023judging} for ASR evaluation and demonstrate that modern LLMs can achieve strong agreement with human judgments when assessing the semantic consistency between ASR outputs and reference transcriptions. This motivates our proposed \textbf{Sentence-level Semantic Error Rate ($S^2$ER)} as a complementary metric to WER to evaluate semantic coherence. In addition, we design an LLM-driven agentic framework that enables ASR systems to interact with users and leverage feedback for iteratively correcting recognition errors. 
Our key contributions are summarized as follows:

\begin{itemize}
    \item \textbf{New ASR Semantic Metric ($S^2ER$):} We propose $S^2ER$, a novel evaluation metric that leverages LLMs as judges to assess ASR semantic success. Through human evaluation, we demonstrate a strong correlation between LLM judgments and human preferences in terms of semantic consistency.
    \item \textbf{Agentic Correction Framework:} We develop an LLM-simulated interactive correction framework to iteratively improve ASR performance.
    \item \textbf{Robust Generalization and Validation:} Extensive experimental validation across diverse linguistic standard benchmarks, demonstrates the framework's broad applicability.
\end{itemize}

\section{Related Works}

While WER has long served as the standard metric for ASR evaluation, its inherent design of assigning equal weight to all words fails to capture critical semantic errors. To address this, several semantic-aware metrics have been proposed. \textbf{Semantic WER} \cite{roy2021semantic} introduced dynamic weighting, which utilizes Named Entity Recognition to extract keywords, assigning higher weights to critical entities and lower weights to filler words during error calculation. Moving to embedding-based evaluation, \textbf{SemDist} \cite{kim2021semantic} utilized RoBERTa-based sentence embeddings to measure semantic similarity beyond literal overlap. Most recently, \textbf{LASER} \cite{parulekar2025laser} leveraged LLMs to assign graded penalties based on error severity (e.g., ignoring colloquial variations while penalizing meaning changes). Unlike these continuous scoring metrics, our LLM-as-a-Judge adopts a binary functional criterion, acting as a strict gatekeeper to determine if the user's intent is executable.

Regarding error correction based on human feedback, traditional ASR systems have explored several approaches, but they often rely on rigid interactions. Early methods utilized \textbf{Multimodal Interfaces} \cite{suhm2001multimodal}, requiring users to manually select alternatives from an N-best list via a GUI or keyboard, which disrupts the hands-free nature of voice interactions. Another approach is \textbf{Acoustic Respeaking} \cite{sperber2013efficient}, where users simply repeat the misrecognized utterance, which is highly inefficient. In contrast, the NLP community has successfully leveraged natural language feedback for error resolution. For instance, agentic frameworks like \textbf{ReAct} \cite{yao2022react} enable LLMs to iteratively reason and refine their outputs based on human or environmental feedback. Inspired by these advancements, our work introduces this interactive paradigm to ASR. Instead of relying on manual edits or rigid repetition, our framework allows users to correct recognition errors using natural, spoken instructions.

\section{Proposed Paradigm}
\label{sec:ProposedParadigm}

Let $I$ denote the user speech and $Y$ the output transcript.
Existing ASR systems operate under a single-pass decoding paradigm:
\begin{equation}
    Y = \text{ASR}(I)
\end{equation}
This formulation is static: once a transcription is produced, the system has no mechanism to incorporate subsequent user feedback. 
\begin{figure}
    \centering
    \centerline{\includegraphics[width=0.85\linewidth]{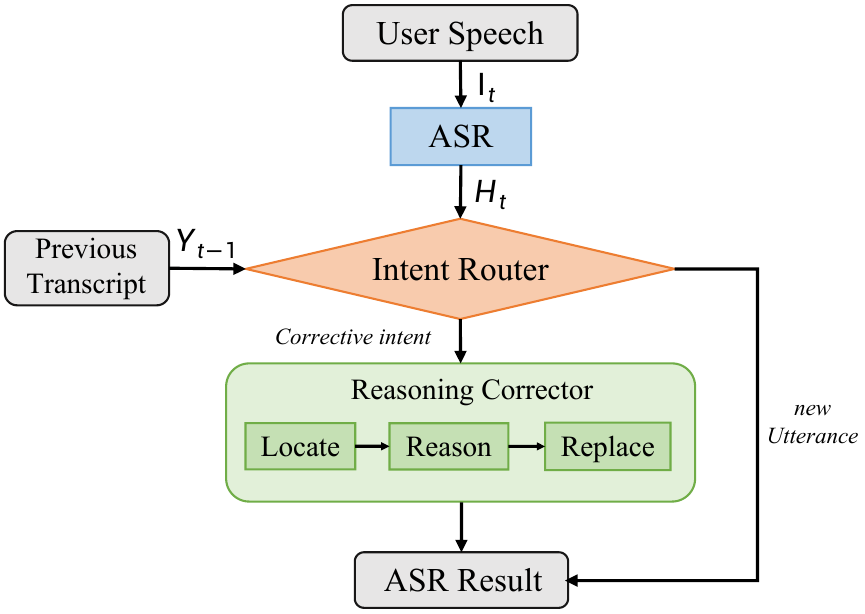}}
    \caption{Overview of the Interactive ASR framework. An LLM Intent Router classifies the base ASR hypothesis ($H_t$) using the previous transcript ($Y_{t-1}$). New utterances bypass correction and are output directly, while corrective instructions trigger an LLM Reasoning Corrector to refine $Y_{t-1}$ via a three-step CoT process (Locate, Reason, Surgical Replacement).}
    \label{fig:Interactive_ASR_Framework}
    \vspace{-0.4cm}
\end{figure}

To address this limitation, we propose  the \textbf{Interactive ASR} framework 
(illustrated in Figure~\ref{fig:Interactive_ASR_Framework}). It cascades a base 
ASR model with an LLM-based reasoning module, enabling the system to refine its output based on user feedback.

At turn $t$, the user provides an input speech $I_t$, which is first 
transcribed by the base ASR model into a text hypothesis $H_t$. The text hypothesis, together with the previous transcript $Y_{t-1}$, is passed to an LLM-based Intent Router, which analyzes the semantic relationship between them to dynamically route the processing pipeline. If $H_t$ does not express any corrective intent toward $Y_{t-1}$, $H_t$ is classified as a new utterance and is directly adopted as the final ASR result $Y_t$. Otherwise, if $H_t$ carries a \textit{Corrective Intent} targeting $Y_{t-1}$, an LLM-based Reasoning Corrector is invoked to update the hypothesis:
\begin{equation}
    Y_t = \text{ReasoningCorrector}(Y_{t-1},\, H_t ;\mathcal{P}_{\text{refine}})
\end{equation}
Guided by a structured prompt $\mathcal{P}_{refine}$, the Reasoning Corrector employs a Chain-of-Thought (CoT) \cite{cot2022wei} approach to systematically refine the transcription through three specific steps:
(1) \textbf{Locate} errors in $Y_{t-1}$ via instruction $H_t$; (2) \textbf{Reason} the intended correction using phonetic or lexical constraints; and (3) \textbf{Surgical Replacement} of erroneous segments while preserving the rest of the sentence.

\section{Automated Simulation Framework}
\label{sec:ASF}

\begin{figure}[htb]
    \centering
    \vspace{-0.3cm}
    \centerline{\includegraphics[width=8cm]{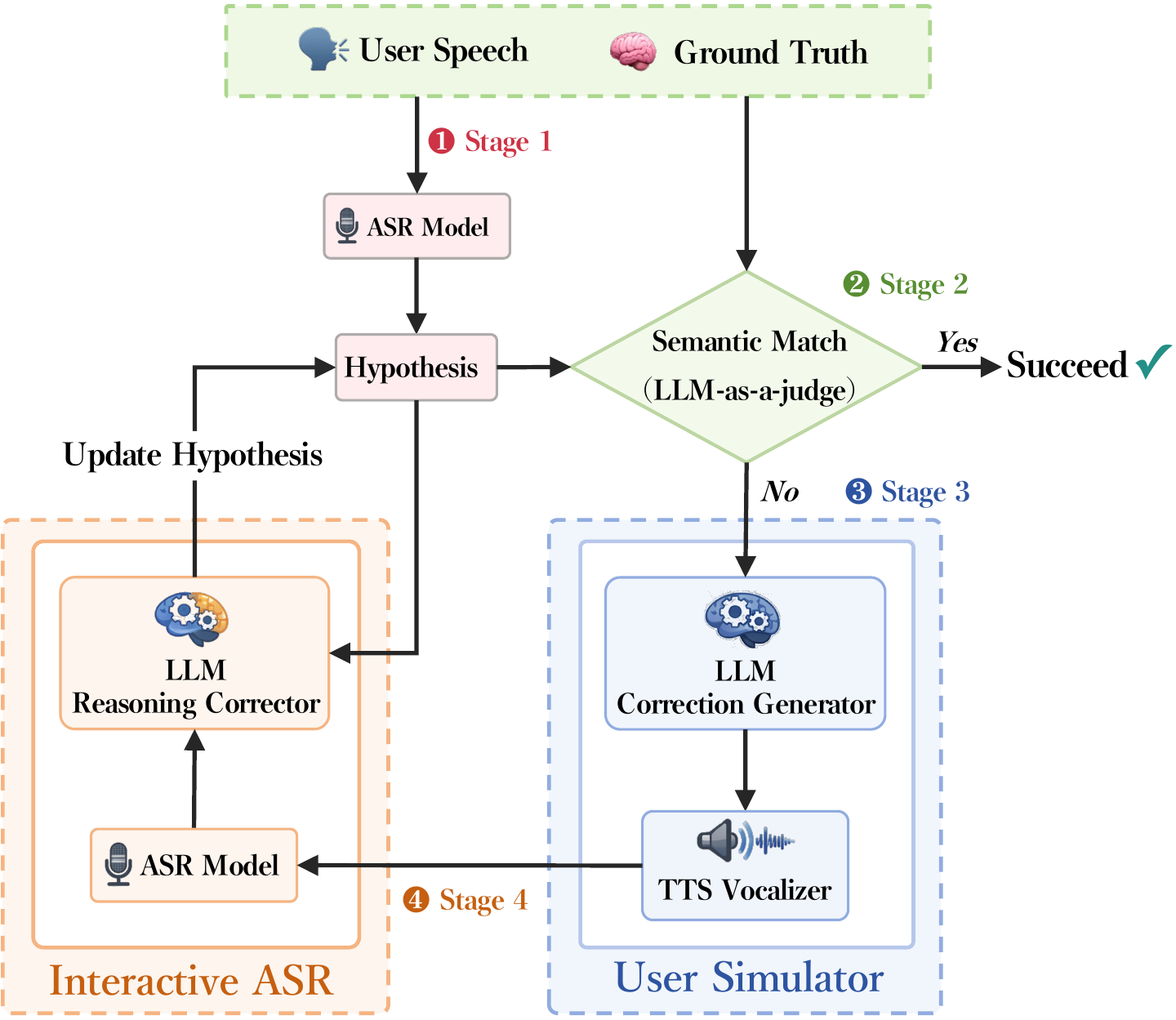}}
    \caption{Overview of the Automated Simulation Framework. An LLM-as-a-judge first evaluates the semantic coherence of the initial ASR hypothesis. Upon detecting an error, the User Simulator generates a spoken correction instruction via an LLM and TTS. The Interactive ASR then processes this feedback to reason and update the hypothesis, forming an automated correction process.}
    \label{fig:Automated_Simulation_Framework}
    \vspace{-0.6cm}
\end{figure}

Large-scale human evaluation of interactive ASR in continuous scenarios is expensive and limits reproducibility. To address this and systematically evaluate the corrective capability of our system (the right branch in Figure \ref{fig:Interactive_ASR_Framework}), we design an automated simulation framework tailored for single-utterance iterative correction. As illustrated in Figure \ref{fig:Automated_Simulation_Framework}, it comprises a User Simulator, the interactive correction modules of our ASR system, and a semantic evaluation module.

\subsection{User Simulator}

The User Simulator acts as an oracle to emulate human corrective behavior. Given a ground-truth transcript $Y_{GT}$, the simulator evaluates the hypothesis $Y_{t-1}$ from the previous turn and generates feedback:

\begin{itemize}
    \item \textbf{Correction Generator:} If a semantic mismatch exists between $Y_{t-1}$ and $Y_{GT}$, the simulator generates a natural language correction $C_t$ conditioned on given prompt $P_{user}$:
    
    \begin{equation}
        C_t = \text{LLM}_{user}(Y_{GT}, Y_{t-1};\mathcal{P}_{user})
    \end{equation}
    
    To ensure the interaction mirrors real-world behavior, $\mathcal{P}_{\text{user}}$ contains diverse human-like strategies, including phonetic spelling, contextual clarification, and direct negation.

    \item \textbf{TTS Vocalizer:} To maintain speaker consistency across turns, we employ a zero-shot voice-cloning TTS model. By utilizing the original source audio $I_0$ as an acoustic reference, the correction text $C_t$ is synthesized into a speech instruction $I_t$:
    
    \begin{equation}
        I_t = \text{TTS}(C_t, I_0)
    \end{equation}
\end{itemize}
\subsection{Interactive ASR System}
In this simulation context, the Interactive ASR system operates as an iterative state-updating engine for a single utterance. The process is initialized at $t=0$, where the base ASR model decodes the original user speech $I_0$ into the initial hypothesis $Y_0$. For each subsequent turn $t > 0$, the system receives a corrective speech $I_t$ from the user simulator. Since the input is strictly a correction in this testing setup, the Intent Router is bypassed. The corrective speech is firstly transcribed by the base ASR model into a text hypothesis $H_t$. Subsequently, the \textbf{Reasoning Corrector} , directly processes the previous state $Y_{t-1}$ alongside $H_t$. By leveraging the same Chain-of-Thought (CoT) reasoning described in Section \ref{sec:ProposedParadigm}, it identifies erroneous segments and performs surgical edits to produce the updated state $Y_t$:
\begin{equation}
    Y_t = \text{ReasoningCorrector}(Y_{t-1}, H_t; \mathcal{P}_{refine})
\end{equation}
This iterative refinement loop continues until the \textbf{Semantic Judge} verifies the transcript against the ground truth or a predefined maximum turn limit is reached.

\subsection{Sentence-level Semantic Error Rate ($S^2ER$)}
\label{ssec:SemanticJudge}
To precisely quantify task-oriented success, we define $S^2ER$ as the average semantic mismatch rate across $N$ utterances:
\begin{equation}
    S^2ER = \frac{1}{N} \sum_{i=1}^{N} \left( 1 - \text{LLM}_{judge}(Y_i, Y_{GT,i}; \mathcal{P}_{judge}) \right)
\end{equation}
where $\text{LLM}_{judge}(\cdot) \in \{0, 1\}$ outputs 1 for semantic equivalence. To ensure $S^2ER$ reflects functional correctness, $\mathcal{P}_{judge}$ instructs the judge to prioritize core intent and critical entities, ignoring minor surface-level variations (e.g., filler words or punctuation).

\section{Experiments}

In this section, we comprehensively evaluate the proposed Interactive ASR framework. We first outline the experimental setup in Section \ref{ssec:ExperimentsSetup}, detailing the diverse benchmarks and the foundational models. Before analyzing the system performance, we conduct a \textbf{Human-AI Alignment Study} in Section \ref{ssec:human_ai_alignment} to establish the credibility of $S^2ER$ by demonstrating its consistency with human intuition. Subsequently, we present the \textbf{Main Results} of the interactive framework in Section \ref{ssec:Mainresult}, utilizing a combination of performance tables, trend curves, and case studies to provide a holistic demonstration of the system's corrective capabilities.

\subsection{Experiments Setup}
\label{ssec:ExperimentsSetup}
\label{sssec:Datasets}

To comprehensively evaluate the robustness and generalization of our Interactive ASR framework across diverse and challenging scenarios, we conduct experiments on three representative benchmarks: \textbf{ASRU2019 Test} \cite{shi2020asru}, a 20-hour test set targeting complex intra-sentential Mandarin-English code-switching; \textbf{GigaSpeech Test} \cite{chen2021gigaspeech}, a 40-hour multi-domain English subset from podcasts and YouTube representing diverse acoustic environments; and \textbf{WenetSpeech Net} \cite{zhang2022wenetspeech}, a 23-hour Mandarin test split of internet-sourced spontaneous speech.
In our experiments, we employ \textbf{Qwen3-ASR-1.7B} \cite{qwen3asr_shi} as the foundational ASR model to generate initial text hypotheses across English, Mandarin, and code-switching scenarios. For cognitive processing, \textbf{Qwen3-32B} \cite{yang2025qwen3} powers both the \textbf{Correction Generator} module within the User Simulator, the \textbf{Reasoning Corrector} in the Interactive ASR, and the 
\textbf{Semantic Judge}. Finally, \textbf{Index-TTS-1.5} \cite{deng2025indextts} serves as the \textbf{TTS Vocalizer} in the User Simulator, utilizing the original speech as an acoustic prompt to maintain timbre consistency across multi-turn interactions.

\subsection{Human-AI Alignment Study}
\label{ssec:human_ai_alignment}
To validate $S^2ER$ as a reliable ASR metric, we investigate its alignment with human perception. Drawing inspiration from established testing paradigms in Spoken Language Assessment \cite{zechner2019automated}, we curated a balanced evaluation subset of 120 ASR hypothesis-ground truth pairs from the three datasets detailed in Section~\ref{sssec:Datasets} (40 pairs per dataset, evenly split between semantically equivalent and non-equivalent instances based on initial LLM predictions). 23 nonprofessional annotators and 5 in-domain experts performed binary judgments ($1$ for semantically equivalent, $0$ for non-equivalent) on these pairs and the averaged rating are served as the human consensus. We then utilized the Pearson correlation coefficient ($r$)\cite{pearson1895} to quantify how closely the automated LLM judge and individual domain experts align with the human consensus.

\begin{table}[htb]
  \centering
  \small 
  \vspace{-0.2cm}
  \caption{Pearson correlation coefficients of LLM and Expert judgments against the Human Ground Truth. \textbf{Bold} values indicate the higher correlation in each row.}
  \vspace{-0.2cm} 
  \label{tab:correlation_human_llm}
  \renewcommand{\arraystretch}{0.85}
  \setlength{\tabcolsep}{4pt} 
  \begin{tabular}{l c c}
    \toprule
    \textbf{Dataset} & \textbf{LLM ($r$)} & \textbf{Expert ($r$)} \\
    \midrule
    GigaSpeech & \textbf{0.8730} & 0.8345 \\
    WenetSpeech & \textbf{0.7873} & 0.7351 \\
    ASRU2019 & 0.8556 & \textbf{0.8613} \\
    \midrule
    \textbf{Overall} & \textbf{0.8281} & 0.8104 \\ 
    \bottomrule
  \end{tabular}
  \vspace{-0.2cm} 
\end{table}

The comparative alignment results are presented in Table~\ref{tab:correlation_human_llm}. Both the LLM and expert evaluations demonstrate robust consistency with overall human judgment. Notably, the LLM judge achieves an overall Pearson coefficient of $0.8281$, effectively surpassing the average alignment of domain experts ($r=0.8104$). These findings explicitly confirm that our LLM-based evaluator exhibits a high degree of fidelity to human semantic perception, firmly establishing its validity as the scoring engine for $S^2ER$.

\subsection{Main results: Multi-turn iteractive performance}
\label{ssec:Mainresult}

\begin{figure}
  \centering
  \includegraphics[width=0.45\textwidth]{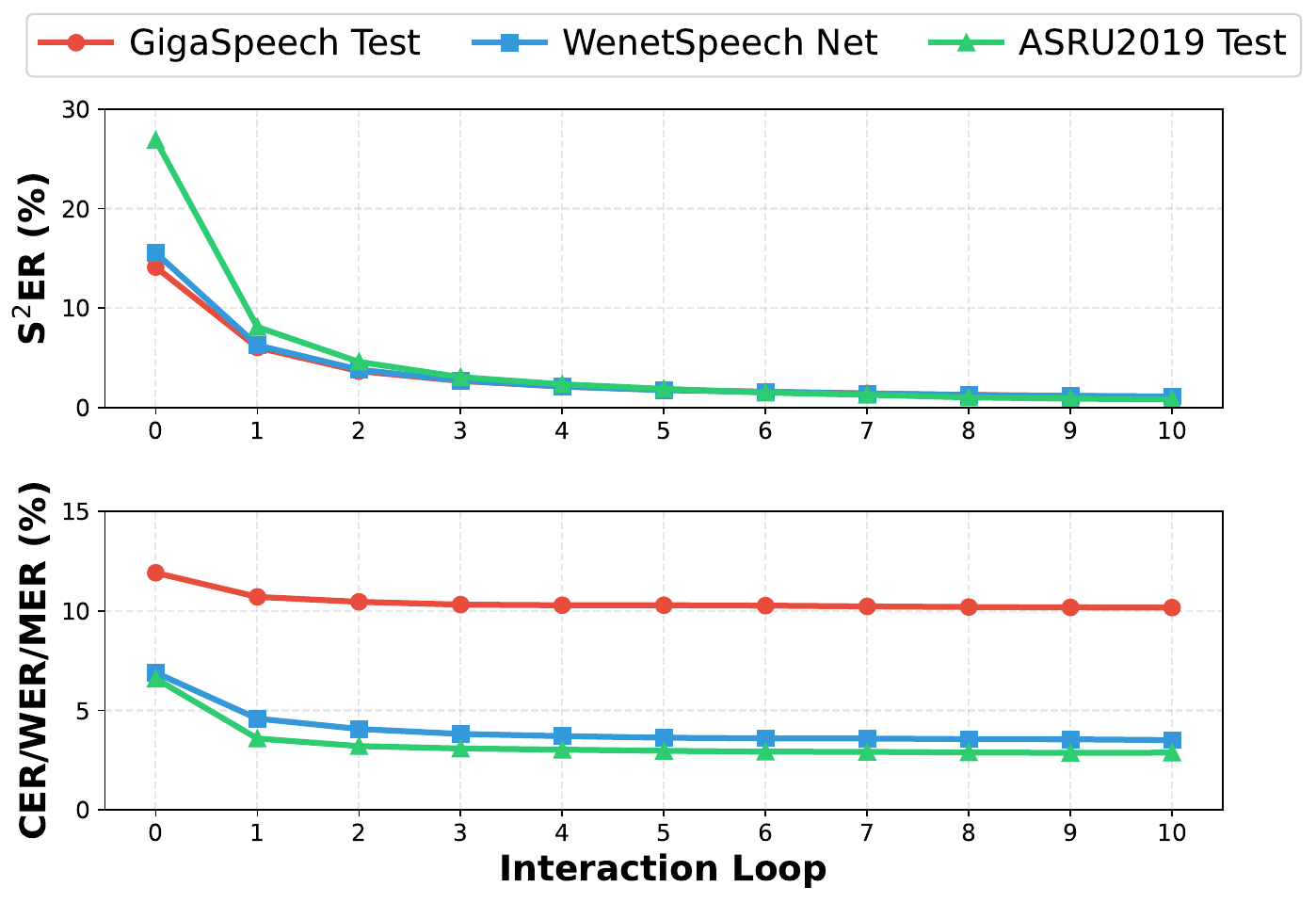}
  \vspace{-0.3cm}
  \caption{Comparison of $S^2ER$ (top) and CER/WER/MER (bottom) reduction trends across three datasets.}
  \label{fig:s2er_wer_trend}
  \vspace{-0.4cm} 
\end{figure}

Table~\ref{tab:main_results} and Figure~\ref{fig:s2er_wer_trend} present the performance of our Interactive ASR framework across three datasets, tracking traditional metrics alongside our proposed $S^2ER$. Specifically, Table~\ref{tab:main_results} reports the exact metric values at key discrete stages (loops 0, 1, 2, 3, and 10), while Figure~\ref{fig:s2er_wer_trend} illustrates the continuous performance trajectory across all 10 interaction loops. 

Crucially, the trend curves reveal a steep improvement concentrated within the initial turns. After just a single interaction loop, $S^2ER$ experiences a dramatic drop: from 14.12\% to 6.03\% on GigaSpeech Test, 15.56\% to 6.26\% on WenetSpeech Net, and 26.89\% to 8.10\% on ASRU2019 Test. By the second loop, $S^2ER$ further decrease to 3.66\%, 3.81\%, and 4.59\%, respectively. In natural human-computer interaction, user patience degrades sharply with repeated failures. By resolving core ambiguities in just 1–2 turns, our framework mirrors natural conversational repair, ensuring a frictionless user experience.

Beyond the second turn, the system continues to yield steady improvements. Although the reduction margin naturally narrows compared to the massive drops in the first two loops, the ongoing gains are still substantial. We report the 10-turn limits (achieving final $S^2ER$s of 1.08\%, 1.11\%, and 0.82\%, respectively) not as practical operational targets, but to establish the system's theoretical upper bound. By this 10th loop, our system achieves near-perfect performance. Qualitative analysis reveals that the very few bad cases at this stage primarily stem from cascading ASR errors during the interaction. When the base ASR repeatedly misrecognizes the user's corrective instructions, the LLM loses the reliable anchors required for surgical replacement, ultimately causing the correction loop to stall.

\begin{table}[tb]
\centering                                                                                                                                                                                  
\caption{Performance evolution of the proposed sentence-level semantic error rate ($S^2$ER) versus conventional metrics (WER, CER, MER, SER) across iterative interaction rounds on three datasets.}                                                                                 
\vspace{-0.35cm}
\label{tab:main_results}                                                                                                                                                                    
\resizebox{\columnwidth}{!}{
\small
\renewcommand{\arraystretch}{0.85}
\setlength{\tabcolsep}{2pt}
\begin{tabular}{l  ccc  ccc  ccc}
  \toprule
  \multirow{2}{*}{\textbf{Loop}} &
  \multicolumn{3}{c}{\textbf{GigaSpeech}} &
  \multicolumn{3}{c}{\textbf{WenetSpeech}} &
  \multicolumn{3}{c}{\textbf{ASRU2019}} \\[1ex]
  \cmidrule(lr){2-4} \cmidrule(lr){5-7} \cmidrule(lr){8-10}
  & WER & SER & $S^2ER$ & CER & SER & $S^2ER$ & MER & SER & $S^2ER$ \\
  \midrule
  0  & 12.25 & 61.17 & 14.12 & 6.89 & 35.24 & 15.56 & 6.60 & 38.85 & 26.89 \\
  1  & 11.08 & 58.56 & 6.03  & 4.59 & 28.59 & 6.26  & 3.59 & 25.22 & 8.10  \\
  2  & 10.82 & 58.03 & 3.66  & 4.07 & 26.97 & 3.81  & 3.21 & 23.04 & 4.59  \\
  3  & 10.68 & 57.80 & 2.67  & 3.82 & 26.30 & 2.71  & 3.09 & 22.08 & 3.06  \\
  10 & 10.53 & 57.59 & 1.08  & 3.51 & 25.32 & 1.11  & 2.88 & 20.88 & 0.82  \\
  \bottomrule
\end{tabular}
}
\vspace{-0.2cm}
\begin{flushleft}
  \footnotesize Note: All values are percentages.  CER = Character Error Rate; WER = Word Error Rate; MER = Mixture error rate, which considers Mandarin characters and English words as the tokens in the edit distance calculation. SER = Sentence Error Rate; $S^2ER$ = Sentence-level Semantic Error Rate
(proposed).
\end{flushleft}
\vspace{-0.4cm}
\end{table}

\section{Conclusion}

In this work, we addressed two critical limitations in traditional ASR: semantic-blind evaluation and the absence of interactive correction mechanisms. We introduced $S^2ER$, a novel metric that leverages LLMs as judges to prioritize sentence-level semantic coherence. Furthermore, we proposed an Interactive ASR framework that employs CoT reasoning to iteratively refine transcripts via spoken feedback.

Our experiments establish $S^2ER$ as a reliable ASR metric, demonstrating a strong correlation with ground-truth semantics that surpasses average human performance. Additionally, our framework proved highly effective across diverse and challenging scenarios, including English, Mandarin, and Mandarin-English code-switching.

While this study highlights the potential of interactive ASR, our current simulations rely on large models to establish an upper performance bound. Future work will explore deploying these frameworks on smaller, constrained architectures, aiming to perfectly balance high-level cognitive reasoning with computational efficiency.

\bibliographystyle{IEEEtran}
\bibliography{mybib}

\end{document}